\documentclass[10pt,journal,compsoc]{IEEEtran}
\ifCLASSOPTIONcompsoc
  \usepackage[nocompress]{cite}
\else
  \usepackage{cite}
\fi

\ifCLASSINFOpdf
 \else
 \fi
 \usepackage{color}
\usepackage{amsmath}
\usepackage{tikz}
\usepackage{array}
\usepackage{times}
\usepackage{subfig}
\usepackage{helvet}
\usepackage{subfig}
\usepackage{amssymb}
\usepackage{amsmath}
\usepackage{amssymb}
\usepackage{courier}
\usepackage{multirow}
\usepackage{mathrsfs}
\usepackage{graphicx}
\usepackage{enumitem}
\usepackage{graphicx}
\usepackage{blindtext}
\usepackage{algorithm}
\usepackage{algorithmic}
\usepackage{lineno}
\usepackage[marginal]{footmisc}
\usepackage{booktabs}
\usepackage{endnotes}
\usepackage{amsthm}
\theoremstyle{plain}
\theoremstyle{definition}
\renewcommand{\paragraph}[1]{\vspace{2ex}\noindent\textbf{#1}.}
\usepackage[utf8]{inputenc}
\usepackage[english]{babel}
\usepackage{epstopdf}
\usepackage{array}
\usepackage{multirow}
\usepackage{booktabs}

\newtheorem{definition}{Definition}

\newcommand{\sysname}{KGE-SymCL}

\def\BibTeX{{\rm B\kern-.05em{\sc i\kern-.025em b}\kern-.08em
    T\kern-.1667em\lower.7ex\hbox{E}\kern-.125emX}}

\ifCLASSOPTIONcompsoc
  \usepackage[nocompress]{cite}
\else
  \usepackage{cite}
\fi

\ifCLASSINFOpdf
\else
\fi

\usepackage{tikz}
\usepackage{mathrsfs}
\usepackage{amssymb}
\usepackage{bm}
\usepackage{amsmath}
\usepackage{multirow}
\usepackage{comment}
\usepackage{ragged2e}
\begin{document}

\title{Knowledge Graph Contrastive Learning based on Relation-Symmetrical Structure}
\author{Ke Liang$^{\ast}$,
        Yue Liu$^{\ast}$,
        Sihang Zhou,
        Wenxuan Tu, Yi Wen, Xihong Yang,\\ Xiangjun Dong$^{\dag}$, Xinwang Liu$^{\dag}$,
        ~\IEEEmembership{Senior~Member,~IEEE},
\IEEEcompsocitemizethanks{\IEEEcompsocthanksitem $^{\ast}$ Equal contribution.
\IEEEcompsocthanksitem $^{\dag}$ Xiangjun Dong and Xinwang Liu are both Corresponding Authors.
\IEEEcompsocthanksitem Ke Liang, Yue Liu, Wenxuan Tu, Yi Wen, Xihong Yang, and Xinwang Liu are with the School of Computer, National University of Defense Technology, Changsha, Hunan,410073, China. E-mail: {xinwangliu@nudt.edu.cn}.
\IEEEcompsocthanksitem Sihang Zhou is with the College of Intelligence Science and Technology, National University of Defense Technology, Changsha, Hunan, 410073, China.
\IEEEcompsocthanksitem Xiangjun Dong is with the School of Computer Science and Technology, Qilu University of Technology, Jinan, Shandong, 250316, China.
}
\thanks{Manuscript Accepted June 1, 2023;}}

%
%

\markboth{IEEE Transactions on Knowledge and Data Engineering}
{Shell \MakeLowercase{\textit{et al.}}: Bare Advanced Demo of IEEEtran.cls for IEEE Computer Society Journals}

\IEEEtitleabstractindextext{%
\begin{abstract}
\justifying
Knowledge graph embedding (KGE) aims at learning powerful representations to benefit various artificial intelligence applications. Meanwhile, contrastive learning has been widely leveraged in graph learning as an effective mechanism to enhance the discriminative capacity of the learned representations. However, the complex structures of KG make it hard to construct appropriate contrastive pairs. Only a few attempts have integrated contrastive learning strategies with KGE. But, most of them rely on language models (\textit{e.g.,} Bert) for contrastive pair construction instead of fully mining information underlying the graph structure, hindering expressive ability. Surprisingly, we find that the entities within a relational symmetrical structure are usually similar and correlated. To this end, we propose a knowledge graph contrastive learning framework based on relation-symmetrical structure, \sysname{}, which mines symmetrical structure information in KGs to enhance the discriminative ability of KGE models. Concretely, a plug-and-play approach is proposed by taking entities in the relation-symmetrical positions as positive pairs. Besides, a self-supervised alignment loss is designed to pull together positive pairs. Experimental results on link prediction and entity classification datasets demonstrate that our \sysname{} can be easily adopted to various KGE models for performance improvements. Moreover, extensive experiments show that our model could outperform other state-of-the-art baselines.
\end{abstract}

\begin{IEEEkeywords}
Knowledge Graph Embedding, Self-Supervised Contrastive Learning, Graph Learning, Symmetrical Property.
\end{IEEEkeywords}}

\maketitle

\IEEEdisplaynontitleabstractindextext
\IEEEpeerreviewmaketitle

\ifCLASSOPTIONcompsoc
\IEEEraisesectionheading{\section{Introduction}\label{sec:introduction}}
\else
\section{Introduction}
\label{sec:introduction}
\fi

\IEEEPARstart{K}{nowledge} graphs (KGs), as a graphical representation of human knowledge, benefit many artificial intelligence applications, such as question answering \cite{ren2021lego}, social recommendation ~\cite{wang2018dkn,Recommand1,LKTKDE}, logic reasoning \cite{teru2020inductive,AKGR}, text generation \cite{song-etal-2020-structural, MKA} and code analysis \cite{ABSLearn}. Motivated by their success, researchers have recently focused on developing better knowledge graph embedding (KGE) models to generate high-quality entity and relation representations for performance improvements.

Recent KGE models can be roughly categorized into three types \cite{KG-Survey2,KG-Survey1,gtkdsurvey,kgetkde2} as follows: (1) translational distance models, \textit{e.g.,} TransE \cite{TransE}, {RotaE} \cite{RotaE}, {QuatE} \cite{QuatE}, {DualE} \cite{DualE}, {HAKE} \cite{HAKE}, (2) semantic matching models, \textit{e.g.,} DisMult \cite{DisMult}, {RESCAL} \cite{RESCAL}, {ComplEX} \cite{ComplEX}, {ConvE} \cite{ConvE}, (3) GNN-based models, \textit{e.g.,} RGCN \cite{R-GCN}, KBGAT \cite{KBGAT}, COMPGCN \cite{COMPGCN}. 
\begin{figure}[t]
\centering
\includegraphics[width=0.48\textwidth]{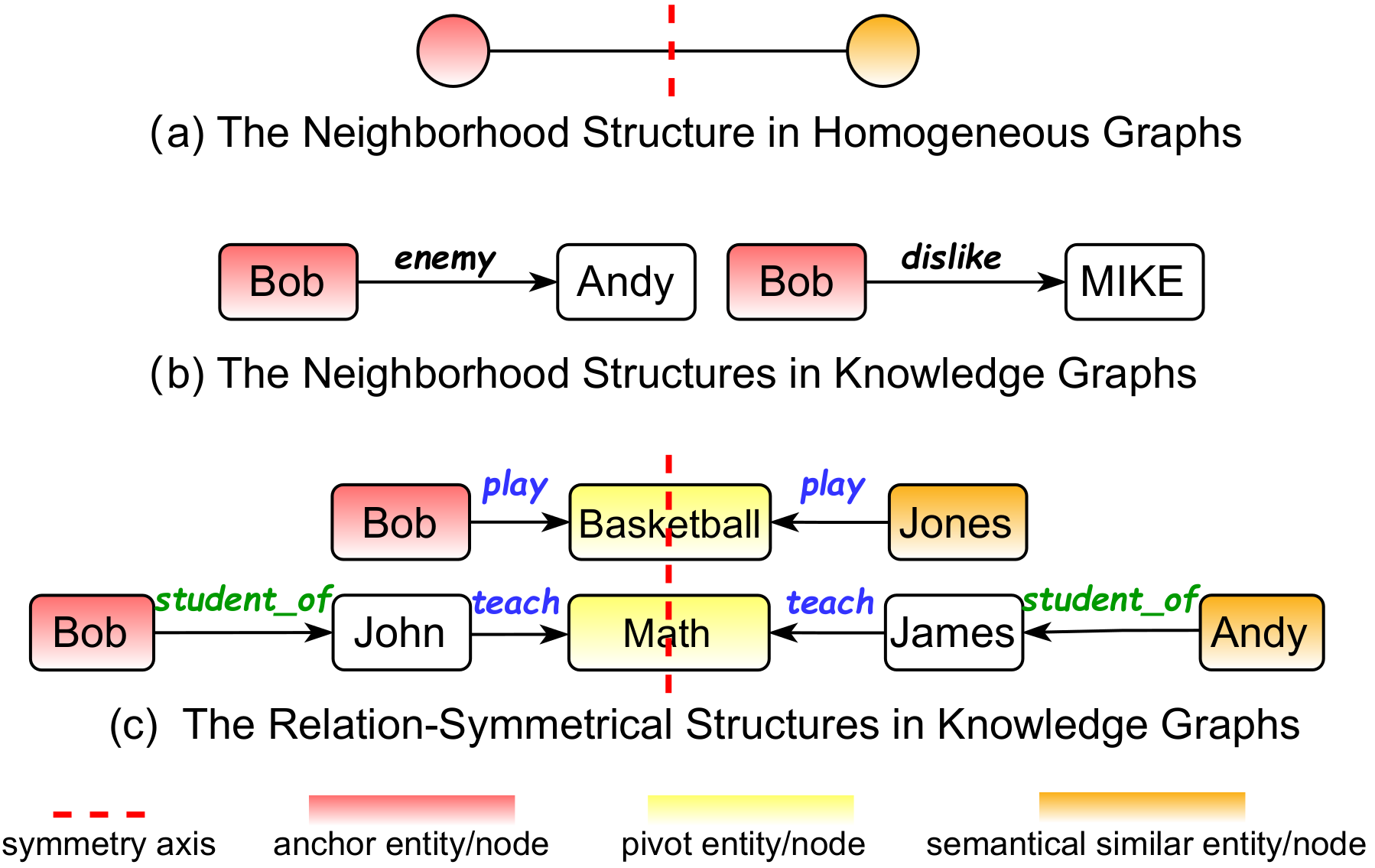} 
\caption{Illustration of the neighborhood and relation-symmetrical structures, where relationships (\textit{i.e.,} \emph{play}, \emph{teach}, \emph{student\_of}) are symmetrical (See Def. \ref{def2}). Sub-Figure (a) and (b) show the differences between neighborhood structures in homogeneous graphs and KGs. The semantics of neighbors may be opposite in KGs, while they are usually assumed to be similar in homogeneous graphs. Sub-Figure (c) shows that symmetrical entities in relation-symmetrical structures will be similar in KGs.}
\label{opposite_example}
\vspace{-4 mm}
\end{figure}
Motivated by the great success of graph contrastive learning, researchers attempt to integrate contrastive learning mechanisms with KGE for more powerful representations \cite{KRACL,simkgc}. Generally speaking, the essence of contrastive learning is to mine the hidden information between samples by pulling together similar samples and pushing away dissimilar samples. Thus, constructing high-confidence contrastive pairs is important to the discriminative ability of contrastive learning models. Until now, only a few works have attempted to integrate contrastive learning strategies with KGE models, such as SimKGC \cite{simkgc}. However, because of the complex structures in the knowledge graphs, these methods usually calculate the semantic similarity via language models (\textit{e.g.,} Bert) to construct the contrastive pairs instead of fully mining information underlying the graph structures like previous graph contrastive learning models.

Although proven effective, the contrastive KGE models in this manner have two apparent limitations. Firstly, the adopted language model can easily affect the performance of these models, \textit{i.e.,} inaccurate contrastive pair construction with inappropriate language models. As a consequence, the performance of these models would drop drastically when language models are not finely trained. Secondly, with the concrete and complicated relationships of entities in the given KG, only relying on the features generated by language models may result in inaccurate semantic estimation. For example, entity \emph{Bob} has opposite semantics with entity \emph{Andy} and \emph{Mike} based on triplets in the given KGs (See Fig. \ref{opposite_example}). However, without considering such structural information, the language models will treat them as similar entities since they all represent human names, which is inaccurate in the given circumstance. The noises caused by such inaccuracy will further hinder the contrastive model for better discriminative ability, which is studied in various works \cite{HSAN, CCGC, zhao2021graph} on graphs.

A more stable and general criterion for contrastive KGE should be developed to solve the problem, such as the graph structures. Neighbors are usually treated to have similar semantics in existing homogeneous graph contrastive learning methods, which benefits the positive contrastive pair construction. However, it is not suitable for knowledge graphs as shown in Fig. \ref{opposite_example} (b). Moreover, we assume that such semantic similarity underlying the neighborhood structures in homogeneous graphs is actually caused by the symmetrical positions of the neighbors. Inspired by it, we observe that the relation-symmetrical structure (See Def. \ref{def2}), which can be commonly found in KGs, will also bring a similar property. This specific structure information will be a good criterion for contrastive KGE. More concretely, entities located in relation-symmetrical positions are usually similar and correlated, and this property can be utilized to construct contrastive positive pairs. For example, the \emph{Bob} and \emph{Jones} are relation-symmetrical about \emph{Basketball} in Fig. \ref{opposite_example} (c), which reveals the similar semantics between \emph{Bob} and \emph{Jones} (i.e., both playing basketball). The observed property is overlooked by the existing contrastive KGE models, thus leading them to sub-optimal performance. 

Following the above idea, we propose a knowledge graph contrastive learning framework based on relation-symmetrical structures, termed \sysname{}. It leverages the symmetrical structural information to enhance the discriminative ability of KGE models. Concretely, a novel plug-and-play approach is designed by taking the entities in the relation-symmetrical positions as positive pairs. Besides, a self-supervised alignment loss is designed to pull together the contrastive positive pairs for more expressive representations. Extensive experimental results on datasets have verified the generalization and effectiveness of our framework. Moreover, the proposed SymCL is easily adopted into the existing KGE models for different downstream tasks. The main contributions of this paper are summarized as follows:
\begin{itemize}
    \item We propose a knowledge graph contrastive learning framework based on relation-symmetrical structures, termed \sysname{}, which is the first work to mine the structural semantics underlying the symmetrical patterns in KGs.

    \item We define Relation-Symmetrical Structures based on the observed symmetry patterns, where positive samples with similar semantics can be naturally found. Besides, we further design a plug-and-play strategy for positive contrastive pair construction, \textit{i.e.,} the entities in the relation-symmetrical positions are treated as the positive pairs.
    
    \item We integrate our \sysname{} with typical KGE baselines and conduct experiments for both link prediction and entity classification. The promising performances verify the effectiveness and generalization of the proposed framework. Moreover, we also compare our \sysname{} to other KGE models, demonstrating the superiority of our approach.
\end{itemize}

The rest parts of the paper are well organized as follows: Section 2 summarizes the related works. Section 3 comprehensively introduces our methodology, termed \sysname{}. The experiments and analysis for our framework from various aspects are presented in Section 4. Section 5 finally concludes the paper.

\section{Related Work}
\subsection{Knowledge Graph Embedding} 
Knowledge Graph Embedding (KGE) aims to encode the entities and relations to the low dimensional vector or matrix space. Recent existing KGE models can be roughly categorized into three types \cite{KG-Survey2,KG-Survey1,TKDE1,TKDE2,TKDE3,MMRNS,GNNORMLP}. (1) Translational distance models leverage distance-based scoring functions and treat the relationship as a translational operation in different latent space, \textit{e.g.,} TransE \cite{TransE}, {RotaE} \cite{RotaE}, {QuatE} \cite{QuatE}, {DualE} \cite{DualE}, {HAKE} \cite{HAKE}, HousE \cite{HousE}, and etc. Among them, TransE \cite{TransE} treats the relation as the addition operation between entities, while {RotaE} \cite{RotaE} regards it as the rotation operation. Moreover, HousE \cite{HousE} involves a novel parameterization based on the designed Householder transformations for rotation and projection. (2) Semantic matching models, including {RESCAL} \cite{RESCAL}, DisMult \cite{DisMult}, ConvE \cite{ConvE}, SimplE \cite{SimplE}, CrossE \cite{CrossE}, QuatE \cite{QuatE}, DualE \cite{DualE}, are developed based on similarity scoring functions. {RESCAL} \cite{RESCAL} first makes use of the bilinear function to associate entities with vectors to capture their latent semantics. Besides, DisMult \cite{DisMult} proposes a multiplication model to represent the likelihood of the triplets. ConvE \cite{ConvE} applies a neural network for similarity modeling. Besides, the advantage of quaternion representations is leveraged by QuatE \cite{QuatE} to enrich the correlation information between head and tail entities based on relational rotation quaternions. Inspired by it, DualE \cite{DualE} is proposed to gain a better expressive ability by projecting the embeddings in dual quaternion space.(3) GNN-based models, including RGCN \cite{R-GCN}, COMPGCN \cite{COMPGCN}, KBGAT \cite{KBGAT}, SCAN \cite{SCAN}, RGHAT \cite{RGHAT} and KE-GCN \cite{KE-GCN}, leverage GNN to capture the structural characteristics of KGs. For example, RGCN \cite{R-GCN} introduces a relation-specific transformation to integrate relation information with message aggregation. RGHAT \cite{RGHAT} is designed with a two-level attention mechanism to handle both relations and entities separately. After that, KE-GCN \cite{KE-GCN} proposes a joint propagation method to update the embedding of nodes and edges simultaneously. COMPGCN \cite{COMPGCN} proposes various composition operations for triplet scoring.
\begin{figure*}[t]
\centering
\includegraphics[width=\textwidth]{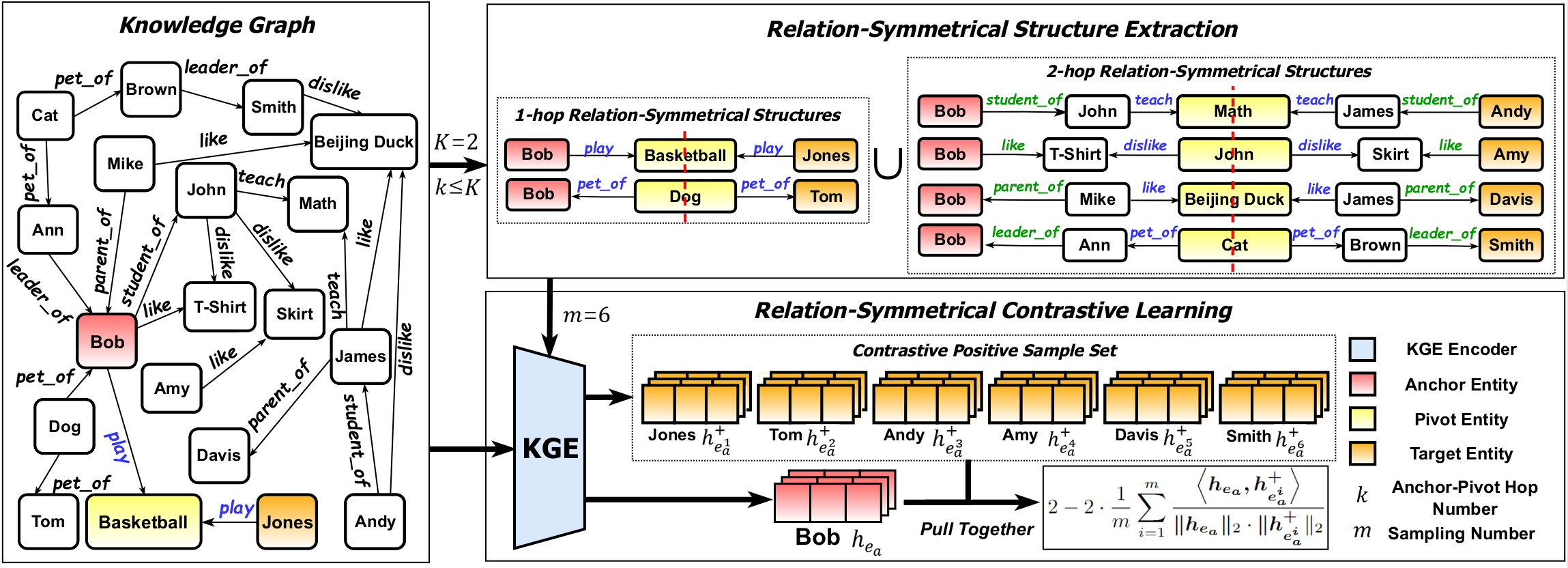} 
\caption{Framework illustration of the proposed \sysname{}. The core idea of \sysname{} is to leverage the semantic similarity of entities in relation-symmetrical positions for constructing contrastive positive samples. The proposed algorithm constitutes two components: relation-symmetrical structure extraction and relation-symmetrical contrastive learning. Concretely, in the relation-symmetrical structure extraction component, we extract the relation-symmetrical structures for the anchor entity Bob and find its target entities, i.e., the positive sample candidates. Then, in the relation-symmetrical contrastive learning component, the entities are embedded by the selected KGE encoder in the latent space. Finally, the designed self-supervised alignment loss guides the network to pull together the positive sample pairs, thus improving the discriminative capability of samples.} 
\label{OVERRALL_FIGURE}  
\vspace{-3 mm}
\end{figure*}

\subsection{Graph Contrastive Learning}
Contrastive Learning (CL), aiming at mining the hidden information in intra-data in a self-supervised manner, has achieved great success in many important fields like natural language processing \cite{NLP_1,NLP_2,fagclTKDE}, computer vision \cite{SIMCLR,BYOL,SIMSAIM,BARLOW,Liliang}, and etc. More recently, Graph Contrastive Learning (GCL) has been a fast-growing field among these research directions. The early works \cite{DGI,MVGRL} demonstrate the effectiveness of the mutual information maximization principle \cite{DIM, contras} in the node and graph level tasks. After that, GRACE \cite{GRACE} and GraphCL \cite{GraphCL} are proposed to pull together the same samples across augmented views and push away the others. However, the large number of negative samples leads to high computational and memory costs. To solve these issues, researchers propose various negative-sample-free methods by redundancy reduction principles \cite{DCRN,IDCRN,G-BT} and asymmetrical strategies \cite{BRGL,AFGRL}. Although verified effectiveness, the promising performance of previous works highly depends on the choice of data augmentation schemes, leading to cumbersome manual trial-and-error. In order to alleviate this problem, the learnable augmentation methods \cite{JOAO,ADGCL}, such as AutoGCL \cite{AutoGCL} and iGCL \cite{iGCL}, are increasingly proposed to generate different views automatically. Also, GCA \cite{GCA} demonstrates the effectiveness of adaptive data augmentations. In addition, the augmentations-free methods are also designed to replace augmentations by discovering the local structural and the global semantics information \cite{AFGRL, liumeng2022embtemporal}, developing parameter un-shared encoders \cite{SCGC}, or perturbing one of the encoders \cite{Simgrace}. More recently, hard sample mining \cite{GCL_study,GCLTKDE} has become another interesting research topic in GCL. To be specific, GDCL \cite{GDCL} utilizes the clustering pseudo labels to correct the bias of the negative sample selection in the attribute graph clustering task. ProGCL \cite{ProGCL} builds a more suitable criterion to handle the hardness of negative samples together with similarity by a designed probability estimator. 

\subsection{Contrastive Learning on Knowledge Graph}
Inspired by the success of graph contrastive learning (See Section 2.2), only a few contrastive KGE models are proposed but are still in the early stage. The existing models, such as SimKGC \cite{simkgc}, construct the contrastive pairs by calculating the semantic similarity estimated by language models, such as Bert \cite{BERT}, etc. Samples with high semantic similarity tend to be combined as positive pairs. 
Although proven effective, the contrastive KGE models in this manner have two apparent limitations. Firstly, the adopted language model can easily affect the performance of these models, \textit{i.e.,} inaccurate contrastive pair construction with inappropriate language models. As a consequence, the performance of these models would drop drastically when language models are not finely trained. Secondly, with the concrete and complicated relationships of entities in the given KG, only relying on the features generated by language models may result in inaccurate semantic estimation. For example, entity \emph{Bob} has opposite semantics with entity \emph{Andy} and \emph{Mike} based on triplets in the given KGs (See Fig. \ref{opposite_example}). However, without considering such structural information, the language models will treat them as similar entities since they all represent human names, which is inaccurate in the given circumstance. The noises caused by such inaccuracy will further hinder the contrastive model for better discriminative ability.
{Previous works \cite{HSAN, CCGC, zhao2021graph,kalantidis2020hard, chu2021cuco, chuang2020debiased, robinson2020contrastive} on graphs have already verified this phenomenon.}
Comparatively, the structural semantics, as more stable semantics underlying all KGs, are rarely used in the existing attempts at integrating KGE and CL due to the complex structures of KGs. The models, such as KGCL \cite{KGCL}, are developed for specific tasks, like recommendation systems. Thus these models have poor generalization on other tasks. To alleviate the above problems, we propose a novel KG contrastive learning framework, termed \sysname{}, leveraging the \textbf{symmetry-structural} semantics, which is also easily adopted to other KGE models and scaled well on various tasks.
\begin{table}[t]
\centering
\fontsize{6}{7}\selectfont 
\caption{Notation summary}
\resizebox{\linewidth}{!}{
\begin{tabular}{cc}
\hline
{Notation} & {Explanation} \\\hline
  $\mathcal{G}$      &    Knowledge graph    \\
$\mathcal{E}$ & Entity set in \emph{KG}\\
$\mathcal{R}$ & Relation set in \emph{KG}\\
$e_{\emph{i}}$ & Entity $i$\\
$r_{\emph{t}}$ & Relation $t$\\
\textbf{h$_{\emph{e}}$} & Embedding vector of entity \emph{e}\\
\textbf{h$_{\emph{e}}^{+}$} & Positive embedding vector of entity \emph{e}\\
$\mathcal{G}_{\emph{inv}}$    &  Knowledge graph with all the edges inversed\\ 
$\mathcal{G}_u$    &  Union of the \emph{KG} and corresponding \emph{KG}$_{\emph{inv}}$\\ 
\emph{P}$_{\mathcal{G}}(\cdot)$    &  Path between $e_{\emph{u}}$ and $e_{\emph{v}}$ in the given \emph{KG}\\ 
\emph{RS}$^{i}(\cdot)$    &  Relation sequence along the $i^{\emph{th}}$ \emph{P}$_{\emph{KG}}$\\ 
$\mathcal{F}$   &  Symbol function of the relation\\ 
\emph{RSym}$(e)$   &  Relation symmetrical structure of entity $e$\\ 
$\mathcal{F}$   &  Symbol function of the relation\\ 
$m$   &  Sampling number\\ 
$k$   &  Anchor-Pivot hop number\\ 
$K$ & The upper bounds for $k$ \\
$\mathcal{CP}_m$(\textbf{h$_{e}$}) & Positive sample set for entity $e$\\
$\bm{g}(\cdot)$ & The selected knowledge graph encoder\\\hline
\end{tabular}}
\label{NOTATION_TABLE} 
\end{table}

\section{Method}
In this section, the details of our relation-symmetrical structure based contrastive knowledge graph framework termed \sysname{} are introduced from two aspects, \textit{i.e.,} Relation-Symmetrical Structure Extraction and Relation-Symmetrical Contrastive Learning (See Fig. \ref{OVERRALL_FIGURE}). Before that, we will first introduce the preliminaries for the method.

\subsection{Preliminary}
Knowledge graph (KG) is composed of the fact triplets, denoted as $\mathcal{G}=\{(e_u,r_{t},e_v)\ |\ e_u,e_v \in \mathcal{E}, r_{t} \in \mathcal{R}\}$, where $\mathcal{E}$ is the set of entities ( \textit{i.e.,} nodes), $\mathcal{R}$ is the set of relations (\textit{i.e.,} edge labels), $e_u$ and $e_v$ represent the head and tail entity respectively, and $r_{t}$ is the relation between them. Based on KGs, we define the relation sequence extraction operation as follows, which is important to understand the relation-symmetrical structure. Besides, we also summarize the notations in Tab. \ref{NOTATION_TABLE}.
\begin{definition}
\textbf{Relation Sequence Extraction}. 
\emph{Given the knowledge graph $\mathcal{G}$$=$$\{(e_u,r_{t},e_v)\ |\ e_u,e_v \in \mathcal{E}, r_{t} \in \mathcal{R}\}$ and the corresponding inversed knowledge graph $\mathcal{G}_{inv}$ $=\{(e_v,r_{t},e_u)\ |\ \forall(e_u,r_{t},e_v)\in$ $\mathcal{G}\}$, Relation Sequence Extraction aims to get the relation sequence $RS^{i}(e_u, e_v)$ along the $i^{th}$ path between $e_u$ and $e_v$ on the $\mathcal{G}_u$$=$$\mathcal{G}$ $\cup$ $\mathcal{G}_{inv}$, {iff the $i^{th}$ path $P_{\mathcal{G}_{u}}(e_u, e_v)$ between $e_u$ and $e_v$ exists in the $\mathcal{G}_u$.} 
\begin{equation} 
\begin{aligned}
RS^{i}(e_u, e_v)=
\{\mathcal{F}(r_{e_u,e^i_{{1}}}), \mathcal{F}(r_{e^i_{{1}},e^i_{2}}),\dots, \mathcal{F}(r_{e^i_{n},e_v})\},
\end{aligned}
\end{equation}where $e^{i}_{n}$ represents the $n^{th}$ entity on the $i^{th}$ path, $r_{e^{i}_{a},e^{i}_{b}}$ represents the relation with the head entity ${e^{i}_{a}}$ and tail entity ${e^{i}_{b}}$ in $\mathcal{G}_u$. Besides, $\mathcal{F}$ is the symbol function defined as follows:
\begin{equation} 
\begin{aligned}
\mathcal{F}(r_{e^{i}_{a},e^{i}_{b}})=\left\{
\begin{array}{rcl}
r_{e^{i}_{a},e^{i}_{b}}^{+}       &      & {(e_{a},r_{e^{i}_{a},e^{i}_{b}},e_{b})\in \mathcal{G}}\\\\
r_{e^{i}_{a},e^{i}_{b}}^{-}       &      & {(e_{a},r_{e^{i}_{a},e^{i}_{b}},e_{b})\in \mathcal{G}_{inv}}\\
\end{array} \right.,
\end{aligned}
\end{equation}}
\end{definition}

\subsection{Relation-Symmetrical Structure Extraction}
Many existing graph contrastive learning frameworks \cite{NCL,GRAPHMLP,NHN} construct the positive pairs based on the structure-semantic similarity of the neighborhood information. However, due to the concrete yet complicated edge labels, the neighbors may not have similar semantics, such as the cases in Fig. \ref{opposite_example} (b).

We assume that the more deep-in reason for the success of the neighbor-based graph contrastive learning frameworks is that they find out the nodes with symmetrical positions based on symmetrical structures in homogeneous graphs as shown in Fig. \ref{opposite_example} (a). Surprisingly, we observe that although such symmetrical similarities between entities may not exist between neighbors in KGs, they can still be found in the relation-symmetrical structures defined in Def. \ref{def2}, where the relationships associated with the edge directions are symmetrical. As shown in Fig. \ref{opposite_example} (c), \emph{Bob} and \emph{Jones} have similar semantics since there are relation-symmetrical patterns between them, \textit{i.e.,} they both play \emph{Basketball} (likewise for \emph{Bob} and \emph{Andy}). In other words, such relation-symmetrical structures will naturally bring positive contrastive pairs with similar semantics. Thus, contrastive pair construction can be reformulated into relation-symmetrical structure extraction in KGs. 
\begin{figure}[t]
\centering
\includegraphics[width=0.48\textwidth]{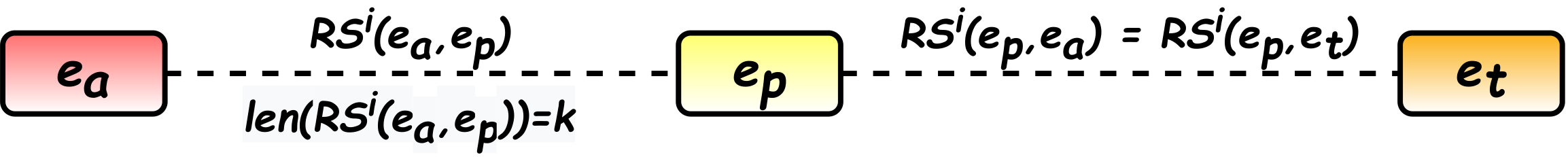}
\caption{{\emph{k-hop} relation-symmetrical structure $RSym^{i}_{k}(e_a)$.}}
\label{definitionofGk}  
\vspace{-0.5 cm}
\end{figure}

\begin{definition}
\textbf{\emph{k-hop} Relation-Symmetrical Structure} \emph{Given the knowledge graph $\mathcal{G}$$=$$\{(e_u,r_{t},e_v)\ |\ e_u,e_v \in \mathcal{E}, r_{t} \in \mathcal{R}\}$, the $i^{th}$ k-hop Relation-Symmetrical Structure of an anchor entity $e_a$ is denoted as $RSym^{i}_{k}(e_a)$, \textit{iff.} the $i^{th}$ structure exists.
\begin{equation} 
\begin{aligned}
RSym^{i}_{k}(e_a)=(\{e_a, e_p, e_t\}, RS^{i}(e_a,e_t)),
\end{aligned}
\end{equation}where $e_p$ is the pivot entity, $e_t$ is the target entity which is symmetrical to $e_a$ about $e_p$. According to $e_p$, the structure can be separated into two parts, where the relation sequence of the first part should be symmetrical to the second part (i.e., $RS^{i}(e_{t}, e_{p})= RS^{i}(e_{a}, e_{p})$). Besides, there are $k$ hops for both two parts, \textit{i.e.,} $len(RS^{i}(e_{a}, e_{p}))=len(RS^{i}(e_{p}, e_{t}))=k$.
\label{def2}}
\end{definition}

Following the ideas above, we propose the relation-symmetrical structure extraction module, which takes the knowledge graph KG, the anchor entity $e_{a}$ and the hyper-parameter $K$ as inputs and outputs the target entity set $\mathcal{P}_{e_{a}}$ for the anchor entity. Concretely, we traverse all of the structures started from the anchor entity in the given KG and only keep the structures satisfying Def. \ref{def2}. Note that the uppercase letter $K$ is the upper bounds for the lowercase letter $k$, \textit{i.e.,} $k\leq K$. Assume the quantity of $k$-$hop$ relation-symmetrical structures for $e_a$ is $n_k$, we can first get the target structure set $\mathcal{T}(e_a,K)=\bigcup_{k\in K}\bigcup_{i\in n_k} RSym_k^{i}(e_a)$. Then, the target entity set $\mathcal{P}_{e_{a}}$ for the anchor entity $e_a$ is generated by picking out all the relation-symmetrical entities in structures belonging to $\mathcal{T}(e_a,K)$. Fig. \ref{OVERRALL_FIGURE} shows an example of the procedure regarding anchor entity \emph{Bob} with $K=2$.

\subsection{Relation-Symmetrical Contrastive Learning} We design a simple yet effective contrastive learning framework to leverage the hidden structural semantics underlying the relation-symmetrical structures to improve the discriminative ability of the KGE models. The entities in the target entity set $\mathcal{P}_{e_{a}}$ are treated as the contrastive positive sample candidates. Based on that, we use a self-supervised alignment loss to pull together the positive pairs for contrastive learning. The details will be illustrated as follows.

\subsubsection{Knowledge Graph Encoding}
Our model can be easily adopted to various KGE models for entity encoding, such as RDF2Vec \cite{RDF2Vec}, RGCN \cite{R-GCN}, COMPGCN \cite{COMPGCN}, {HAKE} \cite{HAKE}, CompLEX-DURA \cite{DURA}, and etc. The selected knowledge graph encoder $\bm{g(\cdot)}$ aims to embed the entity $e$ into the embedding $\textbf{h}_e$ in the latent space.
\begin{equation} 
\begin{aligned}
\textbf{h}_e=\bm{g}(e).
\end{aligned}
\label{loss3}
\end{equation}

\subsubsection{Contrastive Positive Pair Construction} The entities in the target entity set $\mathcal{P}_{e_{a}}$ are treated as the positive candidates for the anchor entity $e_a$. Considering the time efficiency, we random sample $m$ entities within $\mathcal{P}_{e_{a}}$ as the positive samples and feed them into the selected KGE model together with the KG. Therefore, positive pair set $\mathcal{CP}_m(\textbf{h}_{e_a})$ of the anchor entity $e_a$ is generated as follows:
\begin{equation} 
\begin{aligned}
\mathcal{CP}_m(\textbf{h}_{e_a}) = \{(\textbf{h}_{e_a}, \textbf{h}_{{e_{a}^{i}}}^{+})\ |\ {e_{a}^{i}} \in \mathcal{P}_{a}, i\in[1,m] \},
\end{aligned}
\label{loss1}
\end{equation}where $\textbf{h}_{e_{a}^{i}}^{+}$ denotes the embedding of the $i^{th}$ positive sample. Fig. \ref{OVERRALL_FIGURE} shows an example with $m$ set as 6.

\subsubsection{Symmetrical Contrastive Loss}
We design the symmetrical contrastive loss based on a self-supervised alignment loss, \textit{i.e.,} \emph{MSE loss}, used in previous negative-free GCL methods \cite{ermolov2021whitening, BRGL} to pull together the contrastive positive pair $(\textbf{h}_{e_a}, \textbf{h}_{{e_{a}^{i}}}^{+})$ for training:
\begin{equation}
\begin{aligned}
\mathcal{L}_{\emph{contrastive}} &= \frac{1}{m}\sum_{i=1}^{m}{{\emph{MSELoss}}(\bm{\textbf{{h}}_{e_a},\textbf{\emph{h}}_{{e_{a}^{i}}}^{+}})}\\ &= \frac{1}{m}\sum_{i=1}^{m}{\left\| \frac{\bm{\textbf{{h}}_{e_a}}}{\bm{\|\textbf{{h}}_{e_a}\|}_2} - \frac{\bm{\textbf{{h}}_{{e_{a}^{i}}}^{+}}}{\bm{\|\textbf{{h}}_{{e_{a}^{i}}}^{+}\|}_2} \right\|}_2^2 \\&= 2-2 \cdot \frac{1}{m}\sum_{i=1}^{m}{\frac{\left<\bm{\textbf{{h}}_{e_a}, \textbf{{h}}_{{e_{a}^{i}}}^{+}}\right>}{\bm{\|\textbf{{h}}_{e_a}\|}_2 \cdot \bm{\|\textbf{{h}}_{{e_{a}^{i}}}^{+}}\|_2}},
\end{aligned}
\label{positive_loss2}
\end{equation}where $\|\cdot\|_2$ denotes the $\mathcal{L}_2$\emph{-norm}. Our network is optimized by minimizing the contrastive loss. Concretely, the positive samples are pulled together in the latent space in this manner, thus improving the discriminative capability of our network.

\subsubsection{Attributes of \sysname{}}
We emphasize the attributes of KGE-SymCL for a comprehensive understanding. Our contrastive learning framework offers a simple yet effective option to construct contrastive samples only relying on the structure information without using any language models. It is a more stable and lightweight contrastive strategy for KGE. 
\begin{figure}[!t]
\centering
\includegraphics[width=0.48\textwidth]{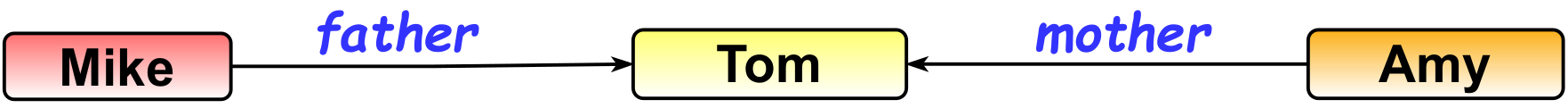} 
\caption{Illustration for entities with similar semantics which are the relational asymmetrical}
\label{beta5}
\end{figure}

Besides, our contrastive framework does not rely on negative samples or graph-augmented views. From the previous observations, the positive samples can be picked up based on the relation-symmetrical structures. However, we observe that it is hard to get high-confident negative constructive pairs only relying on structure information. More concretely, we cannot simply treat the entities beyond those structures as negative samples like previous graph contrastive learning do for the nodes beyond the neighbors. For example, entity \emph{Mike} and \emph{Amy} are not relation-symmetrical in Fig. \ref{beta5}, but we cannot simply regard them as a negative pair. Thus, we design the negative-free contrastive learning framework, which is also a recent trend in graph contrastive learning works \cite{DCRN,IDCRN,BRGL,AFGRL}. Besides, graph augmentations are usually adopted to construct different graph views for contrastive learning. However, the complicated relational structure makes it hard to generate augmented KGs. But, in \sysname{}, the target entities can actually be regarded as another view of the anchor entity in the defined relation-symmetrical structures without any augmentation. This kind of augmentation-free model is also developed in recent graph contrastive learning works \cite{SCGC,AFGRL}. {Moreover, compared to language model (LM) based contrastive KGE models, such as SimKGC \cite{simkgc}, our contrastive sample construction manner is more explainable, since we directly construct contrastive pairs based on the specific attribute in original KGs, instead of the similarity of the embeddings, which can be generated by language models. The latter manner highly relies on whether the LM is well-trained. Due to the deterministic rules used for positive pair construction, our contrastive pairs are more convincing.}

\begin{algorithm}[t]
\fontsize{9}{10.5}\selectfont 
  \caption{Pseudo-code of our \sysname{}}
  \begin{algorithmic}[1]
  \STATE \textbf{Initialization:} Selected KGE encoder: $\bm{g(\cdot)}$; iterations: \emph{T}; target entity set: $\mathcal{P}_{e}$; dictionary \emph{D}$_{\mathcal{P}}$; positive pair set: $\mathcal{CP}_{m}(\cdot)$;
  entity set in \emph{KG}: $\mathcal{E}$; sampling number: $m$; trade-off weight: $\alpha$; 
  \FOR {$e$ in $\mathcal{E}$}
 \STATE Get the target entity set $\mathcal{P}_{e}$ by Eq. (3).
 \STATE Update the \{\emph{e}: $\mathcal{P}_{e}$\} to dictionary \emph{D}$_{\mathcal{P}}$.
  \ENDFOR
  \FOR{\emph{t} = 1 to \emph{T}}
  \FOR {$e$ in $\mathcal{E}$}
  \STATE Generate anchor entity embedding for each triplet with selected KGE model by Eq. (4).
  \STATE Get the target entity set $\mathcal{P}_{e_{a}}$ by checking the generated dictionary \emph{D}$_{\mathcal{P}}$.
  \STATE Generate the contrastive positive embedding pair set with $m$ element $\mathcal{CP}_m(h_{e_a})$ by Eq. (5).
  \STATE Calculate the contrastive loss $\mathcal{L}_{{\emph{contrastive}}}$ by Eq. (6).
  \STATE Optimize the network with Adam by minimizing Eq. (7).
\ENDFOR
\ENDFOR
\end{algorithmic}
\end{algorithm}

\subsection{Training Objective}
The overall training objective of our proposed \sysname{} consists of contrastive loss and task loss. It is formulated as follows:
\begin{equation} 
\begin{aligned}
\mathcal{L} &=\mathcal{L}_{{\emph{task}}} + \alpha \cdot \mathcal{L}_{\emph{contrastive}},
\end{aligned}
\label{loss2}
\end{equation}where $\alpha$ denotes the trade-off hyper-parameter. Our model is adopted for various tasks based on the task-specific loss function, \textit{i.e.,} $\mathcal{L}_{{\emph{task}}}$. For the link prediction, there are many types of loss functions, \textit{e.g.,} the ranking losses \cite{TransE}, the binary logistic regression loss \cite{Ji2016KnowledgeGC}, the sampled multi-class log loss\cite{ComplEX}, the binary cross-entropy loss\cite{COMPGCN}, and etc. Besides, as for entity classification, researchers widely use the cross entropy loss \cite{R-GCN} as the optimization objective. {Note that compared to some of the other contrastive learning frameworks, our \sysname{} is more like a plug-and-play auxiliary module, which should be coupled with the task losses. The labeled data in the supervised loss will help constrain the training procedure to avoid the representational collapse situation. Besides, other common contrastive learning models usually treat the sample $i$ and the augmented sample $i'$ as the positive pair, where the augmented sample is constructed from the original sample. These two samples may be relatively similar in themselves, and it is easier to make the contrastive loss of positive sample pairs tend to 0, resulting in a trivial solution. However, our \sysname{} leverages the symmetrical attributes underlying the data itself, and different samples are treated as positive pairs, where the samples with our contrastive pair construction strategy will naturally have more unique properties inherited from the raw data, thus avoiding the trivial solution in another way.} Algorithm 1 presents the pseudo-code of our \sysname{}.

\section{Experiment and Analysis}
In this section, we first introduce the experiment setup. Then, we integrate our \sysname{} with typical KGE baselines and conduct experiments for link prediction and entity classification to verify the effectiveness and generalization of the proposed framework. Besides, we compare \sysname{}l with other state-of-the-art models to demonstrate its superiority. Afterward, statistical analysis is performed on relation-symmetrical structures, together with intuitive case studies and running time analysis, to comprehensively understand \sysname{}. Finally, we conduct the hyper-parameter experiment to analyze the sensitivity.

\subsection{Experiment Setup}
\subsubsection{Dataset}
For the link prediction task, we evaluate the \sysname{} on three benchmark datasets, \textit{i.e.,} FB15k-237 \cite{FB15K-237}, WN18RR \cite{WN18RR}, and NELL-995 \cite{NELL-995}. For the entity classification task, we adopt four benchmark datasets, including AIFB \cite{AIFB}, AM \cite{AM}, MUTAG \cite{mutag}, and BGS \cite{BGS}. The details of these seven datasets are described below. Besides, Tab. \ref{LP_dataset} and Tab. \ref{EC_dataset} summarize their statistics. As for the train/validation/test splits of these benchmarks, we follow previous works \cite{RGCN-benchmark}.
\begin{itemize}
    \item \textbf{WN18RR}, as a typical link prediction dataset, contains lexical relation triples between words  from WordNet in English. Unlike WN18, there is no inverse relation test leakage in the validation and test datasets in WN18RR. 
    
    \item \textbf{FB15k-237}, as a typical link prediction dataset, are also derived from Freebase, which composes of knowledge base relation triples and textual mentions of their entity pairs. Unlike FB15k,  there is no inverse relation test leakage in the validation and test datasets in FB15k-237. 

    \item \textbf{NELL-995} is derived from the NELL system in the $995^{th}$ iteration. It is designed for link prediction.
    
    \item \textbf{AIFB} describes the AIFB research institute in terms of its research group, publications and staff. It is a Semantic Web (RDF) dataset used as a benchmark in data mining. It is mainly designed for the entity classification task. 
    
    \item \textbf{MUTAG} is a collection of nitroaromatic compounds, aiming at predicting their mutagenicity on Salmonella typhimurium. In this work, we only use it for entity classification. 
    
    \item \textbf{BGS} represents the British Geological Survey as the premier center for expertise and earth science information in the United Kingdom. This data is published as part of BGS OpenGeoscience. We use it for node classification here. 
    
    \item \textbf{AM} describes the cultural heritage objects which correspond to the city of Amsterdam in terms of the museum. It is used for the entity classification task in this paper. 
\end{itemize}
\begin{table}[t]
\fontsize{13}{16}\selectfont 
\caption{Three benchmark datasets for link prediction.}
\resizebox{\linewidth}{!}{
\begin{tabular}{cccccc}
\hline
\multirow{1}{*}{Datasets}  & \multirow{1}{*}{Entities} & \multirow{1}{*}{Relations} & \multirow{1}{*}{Train Edges} & \multirow{1}{*}{Val. Edges} & \multirow{1}{*}{Test Edges} \\
\hline
\multirow{1}{*}{WN18RR}    & \multirow{1}{*}{40,943}   & \multirow{1}{*}{11}        & \multirow{1}{*}{86,835}      & \multirow{1}{*}{3,034}      & \multirow{1}{*}{3,134}      \\
\multirow{1}{*}{FB15K-237} & \multirow{1}{*}{14,541}   & \multirow{1}{*}{237}       & \multirow{1}{*}{272,115}     & \multirow{1}{*}{17,535}     & \multirow{1}{*}{20,466}     \\
\multirow{1}{*}{NELL-995}  & \multirow{1}{*}{75,492}   & \multirow{1}{*}{200}       & \multirow{1}{*}{126,176}     & \multirow{1}{*}{13,912}     & \multirow{1}{*}{14,125}     \\\hline
\end{tabular}}
\label{EC_dataset}
\vspace{8pt}
\fontsize{5}{6.5}\selectfont 
\caption{Four benchmark datasets for entity classification.}
\vspace{-0.2 cm}
\resizebox{\linewidth}{!}{
\begin{tabular}{ccccc}
\hline
\multirow{1}{*}{Datasets} & \multirow{1}{*}{Entities}  & \multirow{1}{*}{Relations} & \multirow{1}{*}{Edges}     &  \multirow{1}{*}{Classes} \\\hline
\multirow{1}{*}{AIFB}     & \multirow{1}{*}{8,285}     & \multirow{1}{*}{45}        & \multirow{1}{*}{29,043}  & \multirow{1}{*}{4}       \\
\multirow{1}{*}{MUTAG}    & \multirow{1}{*}{23,644}    & \multirow{1}{*}{23}        & \multirow{1}{*}{74,227}  & \multirow{1}{*}{2}       \\
\multirow{1}{*}{BGS}      & \multirow{1}{*}{333,845}   & \multirow{1}{*}{103}       & \multirow{1}{*}{916,199}   & \multirow{1}{*}{2}       \\
\multirow{1}{*}{AM}       & \multirow{1}{*}{1,666,764} & \multirow{1}{*}{133}       & \multirow{1}{*}{5,988,321}  & \multirow{1}{*}{11}   \\ \hline
\end{tabular}}
\label{LP_dataset}
\end{table}
\begin{table*}[!t]
\fontsize{8}{11}\selectfont
\caption{Performance comparison between various typical knowledge graph encoders w./w.o. our SymCL framework on link prediction. The boldface values indicate the best results.}
\resizebox{\linewidth}{!}{
\begin{tabular}{ccccccccccccc}
\hline
\multicolumn{1}{c}{\multirow{2.5}{*}{Methods}} & \multicolumn{4}{c}{\multirow{1}{*}{WN18RR}}                                                                                                                & \multicolumn{4}{c}{\multirow{1}{*}{FB15K-237}}                                                                                                                             & \multicolumn{4}{c}{\multirow{1}{*}{NELL-995}}                                                                                                     \\
 \cline{2-13} 
\multicolumn{1}{c}{}                         & \multirow{1}{*}{MRR}               & \multirow{1}{*}{Hit@1}             & \multirow{1}{*}{Hit@3}             & \multicolumn{1}{c}{\multirow{1}{*}{Hit@10}} & \multirow{1}{*}{MRR}               & \multicolumn{1}{c}{\multirow{1}{*}{Hit@1}} & \multicolumn{1}{c}{\multirow{1}{*}{Hit@3}} & \multicolumn{1}{c}{\multirow{1}{*}{Hit@10}} & \multirow{1}{*}{MRR}               & \multirow{1}{*}{Hit@1}             & \multirow{1}{*}{Hit@3}             & \multirow{1}{*}{Hit@10}                                             \\ \hline
\multicolumn{13}{c}{\textit{\textbf{Translational Distance Models}}}                                                                                                                                                                                       \\ \midrule[0.25pt]
\multicolumn{1}{c}{TransE}                   & 0.231                              & 0.021                              & 0.409                              & \multicolumn{1}{c}{0.533}                   & 0.289                              & \multicolumn{1}{c}{0.193}                  & \multicolumn{1}{c}{0.326}                  & \multicolumn{1}{c}{0.478}                   & 0.249                              & 0.095                              & 0.377                              & 0.471                              \\
\multicolumn{1}{c}{TransE-SymCL}                & \multicolumn{1}{c}{\textbf{0.233}} & \multicolumn{1}{c}{\textbf{0.022}} & \multicolumn{1}{c}{\textbf{0.411}} & \multicolumn{1}{c}{\textbf{0.535}}          & \multicolumn{1}{c}{\textbf{0.290}} & \textbf{0.195}                             & {0.326}                             & \multicolumn{1}{c}{\textbf{0.486}}          & \multicolumn{1}{c}{\textbf{0.255}} & \multicolumn{1}{c}{\textbf{0.102}} & \multicolumn{1}{c}{\textbf{0.379}} & \multicolumn{1}{c}{\textbf{0.475}} \\ \hline
\multicolumn{1}{c}{HAKE}                     & {0.497}                     & {0.453}                     & {0.515}                     & \multicolumn{1}{c}{0.582}                   & 0.335                              & \multicolumn{1}{c}{0.237}                  & \multicolumn{1}{c}{0.371}                  & \multicolumn{1}{c}{0.530}                   & 0.415                              & 0.313                              & 0.464                              & 0.612                              \\
\multicolumn{1}{c}{HAKE-SymCL}               & \multicolumn{1}{c}{{0.497}}          & \multicolumn{1}{c}{\textbf{0.454}}          & \multicolumn{1}{c}{{0.515}}       & \multicolumn{1}{c}{\textbf{0.585}}                   & \multicolumn{1}{c}{\textbf{0.346}} & \textbf{0.248}                             & \textbf{0.384}                             & \multicolumn{1}{c}{\textbf{0.544}}          & \multicolumn{1}{c}{\textbf{0.419}} & \multicolumn{1}{c}{\textbf{0.318}} & \multicolumn{1}{c}{\textbf{0.468}} & \multicolumn{1}{c}{\textbf{0.616}} \\ \hline
\multicolumn{13}{c}{\textit{\textbf{Semantic Matching Models}}}                                                                                                                                                                                       \\ \hline
\multicolumn{1}{c}{DisMult}                  & 0.420                              & 0.370                              & 0.439                              & \multicolumn{1}{c}{0.521}                   & 0.243                              & \multicolumn{1}{c}{0.191}                  & \multicolumn{1}{c}{0.271}                  & \multicolumn{1}{c}{0.328}                   & 0.223                              & 0.139                              & 0.244                              & 0.398                              \\
\multicolumn{1}{c}{DisMult-SymCL}            & \multicolumn{1}{c}{\textbf{0.421}} & \multicolumn{1}{c}{\textbf{0.371}} & \multicolumn{1}{c}{\textbf{0.441}} & \multicolumn{1}{c}{\textbf{0.522}}          & \multicolumn{1}{c}{\textbf{0.260}} & \textbf{0.215}                             & \textbf{0.282}                             & \multicolumn{1}{c}{\textbf{0.339}}          & \multicolumn{1}{c}{\textbf{0.224}} & \multicolumn{1}{c}{\textbf{0.140}} & \multicolumn{1}{c}{\textbf{0.248}} & \multicolumn{1}{c}{\textbf{0.401}} \\ \hline
\multicolumn{1}{c}{ComplEX-DURA}             & 0.489                              & 0.445                              & 0.503                              & \multicolumn{1}{c}{0.574}                   & \multicolumn{1}{c}{0.370}          & 0.275                                      & 0.409                                      & \multicolumn{1}{c}{0.562}                   & 0.468                              & 0.375                              & \textbf{0.513}                     & 0.645                              \\
\multicolumn{1}{c}{ComplEX-DURA-SymCL}       & \multicolumn{1}{c}{\textbf{0.491}} & \multicolumn{1}{c}{\textbf{0.448}} & \multicolumn{1}{c}{\textbf{0.504}} & \multicolumn{1}{c}{\textbf{0.576}}          & \multicolumn{1}{c}{\textbf{0.371}} & \textbf{0.276}                             & \textbf{0.411}                             & \multicolumn{1}{c}{\textbf{0.566}}          & \multicolumn{1}{c}{\textbf{0.469}} & \multicolumn{1}{c}{\textbf{0.376}} & \multicolumn{1}{c}{0.512}          & \multicolumn{1}{c}{\textbf{0.647}} \\ \hline
\multicolumn{13}{c}{\textit{\textbf{GNN-based Models}}}                                                                                                                                                                                       \\ \hline
\multicolumn{1}{c}{RGCN}                     & 0.427                              & 0.382                              & 0.446                              & \multicolumn{1}{c}{{0.510}}                   & 0.248                              & \multicolumn{1}{c}{0.153}                  & \multicolumn{1}{c}{0.258}                  & \multicolumn{1}{c}{0.414}                   & 0.382                              & 0.272                              & 0.435                              & 0.590                              \\
\multicolumn{1}{c}{RGCN-SymCL}               & \multicolumn{1}{c}{\textbf{0.432}} & \multicolumn{1}{c}{\textbf{0.396}} & \multicolumn{1}{c}{\textbf{0.447}} & \multicolumn{1}{c}{{0.510}}         & \multicolumn{1}{c}{\textbf{0.249}} & \textbf{0.159}                             & \textbf{0.270}                             & \multicolumn{1}{c}{\textbf{0.435}}          & \multicolumn{1}{c}{\textbf{0.394}} & \multicolumn{1}{c}{\textbf{0.287}} & \multicolumn{1}{c}{\textbf{0.443}} & \multicolumn{1}{c}{\textbf{0.603}} \\ \midrule[0.25pt]
\multicolumn{1}{c}{COMPGCN}                  & 0.469                              & 0.434                              & 0.482                              & \multicolumn{1}{c}{0.537}                   & 0.352                              & \multicolumn{1}{c}{0.261}                  & \multicolumn{1}{c}{0.387}                  & \multicolumn{1}{c}{0.534}                   & 0.456                              & 0.361                              & 0.507                              & 0.637                              \\
\multicolumn{1}{c}{COMPGCN-SymCL}            & \multicolumn{1}{c}{\textbf{0.471}} & \multicolumn{1}{c}{\textbf{0.437}} & \multicolumn{1}{c}{\textbf{0.484}} & \multicolumn{1}{c}{{0.537}}          & \multicolumn{1}{c}{\textbf{0.354}} & \textbf{0.262}                             & \textbf{0.389}                             & \multicolumn{1}{c}{\textbf{0.537}}          & \multicolumn{1}{c}{\textbf{0.469}} & \multicolumn{1}{c}{\textbf{0.378}} & \multicolumn{1}{c}{\textbf{0.520}} & \multicolumn{1}{c}{\textbf{0.649}} \\ \hline
\multicolumn{13}{c}
{\textit{\textbf{Existing Contrastive KGE Models}}}                                                                                                                                                                                   \\
\hline
\multicolumn{1}{c}{SimKGC}                     & 0.652                              & 0.542                             & 0.709                              & \multicolumn{1}{c}{{0.781}}                   & 0.322                              & \multicolumn{1}{c}{\textbf{0.236}}                  & \multicolumn{1}{c}{0.352}                  & \multicolumn{1}{c}{0.501}                   &   -                            &    -                          &       -                        &      -                         \\
\multicolumn{1}{c}{SimKGC-SymCL}               & \multicolumn{1}{c}{\textbf{0.657}} & \multicolumn{1}{c}{\textbf{0.546}} & \multicolumn{1}{c}{{0.709}} & \multicolumn{1}{c}{\textbf{0.791}}         & \multicolumn{1}{c}{\textbf{0.324}} & \multicolumn{1}{c}{0.235}                             & \textbf{0.354}                             & \multicolumn{1}{c}{\textbf{0.504}}          & \multicolumn{1}{c}{{-}} & \multicolumn{1}{c}{{-}} & \multicolumn{1}{c}{{-}} & \multicolumn{1}{c}{{-}} \\
\hline
\end{tabular}}
\label{LP_ADD}
\end{table*}

\subsubsection{Implementation Detail}
We implement \sysname{} based on the PyTorch library \cite{paszke2019pytorch} and conduct all experiments with a single NVIDIA TITAN XP GPU and intel core i9-9900K CPU. For a fair comparison, the parameter settings in \sysname{} for the KGE encoder are the same as shown in the original paper. As for the specific hyper-parameters used in our work, we search the upper bound of the anchor-pivot hop number $K$ in $\{1, 2, 3\}$, the sampling number $m$ in $\{10, 50, 100, 1000\}$, and the trade-off weight $\alpha$ for contrastive loss in $\{0.001, 0.01, 0.1, 1\}$. The best models on each dataset are listed in the result tables. As for the evaluation metrics, we adopt classification accuracy to evaluate the entity classification performance. As for the link prediction, the head or tail entity of a true triplet is substituted with another entity for each triplet in the test set, which is regarded as the candidate triplet. We further deploy the trained model on all candidate triplets and rank them based on the calculated score. Ideally, true triplets should get higher scores than others. Subsequently, following previous works in this domain, the quality of the prediction is evaluated based on the rank-based measures, including Mean Reciprocal Rank (MRR) and Hits@N, where N $\in \{1,3,10\}$. The mean results of three runs of each experiment are reported {\footnote{\small variances about results are very small in all cases, thus not reported}} as previous works do \cite{COMPGCN, simkgc}.

\subsubsection{Compared Baseline} We compare \sysname{} with existing state-of-the-art KGE models. As for the link prediction task, fourteen models are selected as the compared baselines, including {TransE} \cite{TransE}, {RotaE} \cite{RotaE}, {QuatE} \cite{QuatE}, {DualE} \cite{DualE}, {HAKE} \cite{HAKE}, {RESCAL} \cite{RESCAL}, {DisMult} \cite{DisMult}, {ComplEX} \cite{ComplEX}, {ConvE} \cite{ConvE}, {CompLEX-DURA} \cite{DURA}, {RGCN} \cite{R-GCN}, {SCAN} \cite{SCAN}, {KBGAT} \cite{KBGAT}, {COMPGCN} \cite{COMPGCN}. Moreover, we also extend our model with two contrastive KGE learning models, \textit{i.e.,} SimKGC \cite{simkgc}. As for the entity classification task, Feat \cite{FEAT}, {WL} \cite{WL}, {RDF2Vec} \cite{RDF2Vec}, {RR-GCN} \cite{RR-GCN} are selected as the compared baselines. Note that except for the results for NELL-995, the other results of the compared baselines are recorded from the original papers. 
\begin{table}[t]
\fontsize{6}{7.5}\selectfont 
\caption{Performance comparison between various typical knowledge graph encoders w./w.o. our SymCL framework on entity classification. The boldface values indicate the best results.}
\resizebox{\linewidth}{!}{
\begin{tabular}{ccccc}
\hline
{Methods}        & {AIFB}           & {MUTAG}          & {BGS}            & {AM}             \\\hline
\multirow{1}{*}{RDF2Vec}         & \multirow{1}{*}{{88.88}} & \multirow{1}{*}{72.06}          & \multirow{1}{*}{86.21}          & \multirow{1}{*}{87.88}          \\
\multirow{1}{*}{RDF2Vec-SymCL}   & \multirow{1}{*}{{88.88}} & \multirow{1}{*}{\textbf{73.53}} & \multirow{1}{*}{\textbf{89.66}} & \multirow{1}{*}{\textbf{88.89}}
                             \\\hline 
\multirow{1}{*}{RGCN}       & \multirow{1}{*}{95.83}          & \multirow{1}{*}{72.21}          & \multirow{1}{*}{81.38}          & \multirow{1}{*}{89.19}          \\
\multirow{1}{*}{RGCN-SymCL} & \multirow{1}{*}{\textbf{96.11}} & \multirow{1}{*}{\textbf{72.35}} & \multirow{1}{*}{\textbf{83.45}} & \multirow{1}{*}{\textbf{89.60}}
     \\ \hline
\multirow{1}{*}{COMPGCN}   & \multirow{1}{*}{{94.44}} & \multirow{1}{*}{{79.29}} & \multirow{1}{*}{{82.35}} & \multirow{1}{*}{{93.10}}\\
\multirow{1}{*}{COMPGCN-SymCL}   & \multirow{1}{*}{{94.44}} & \multirow{1}{*}{\textbf{80.90}} & \multirow{1}{*}{\textbf{88.24}} & \multirow{1}{*}{\textbf{96.55}}
     \\\hline 
\end{tabular}}
\label{EC_table}
\end{table}

\subsection{Performance Comparison}
\subsubsection{Effectiveness of \sysname{}}
To verify the effectiveness and plug-and-play attribute of \sysname{}, Tab. \ref{LP_ADD} and Tab. \ref{EC_table} show the performance comparison of the KGE models w./w.o. our contrastive learning framework. We select eight baseline models for performance comparison. As for the link prediction task, we choose two typical baselines, \textit{i.e.,} the most classical and the most recent state-of-the-art baselines, for each type of the KGE model. Moreover, we also extend our structural-based contrastive learning to the language-model based contrastive KGE model SimKGC. Tab. \ref{LP_ADD} indicates that \sysname{} makes average boosts on all the metrics for most of the baselines. Referring to performance improvements in this area \cite{HAKE} \cite{COMPGCN}, the performances are promising. In particular, our SymCL framework improves the RGCN and COMPGCN on NELL-995 average of 3.1\%, 3.0\%, 5.2\%, 3.2\% on MRR, Hit@1, Hit@3, and Hit@10 separately. Besides, DisMult and Hake's performance on FB15K-237 is also improved, \textit{i.e.,} 1.4\%, 1.8\%, 1.2\%, 1.3\% on MRR, Hit@1, Hit@3, and Hit@10. Moreover, it further indicates that our approach makes average 0.71\%, 1.0\% improvements on MRR, and Hit@10 on the SimKGC, which verifies that \sysname{} can also be well scaled to other KGE-CL methods, and further suggests that the leveraged structural information could also enhance the discriminative ability of other contrastive learning KGE models. As for the entity classification, we observe that \sysname{} makes an average 1.92\% improvement in accuracy compared to the KGE baselines based on Tab. \ref{EC_table}. Moreover, to verify the statistical significance of our framework, we have also conducted a student's t-test between the typical KGE baselines (\textit{i.e.,} COMPGCN and RGCN) w./w.o. SymCL. Tab. \ref{ttestR} shows that the p-values of the MRR metric are always smaller than 0.05, which indicates that the performance improvement is significant statistically.
\begin{table}[t]
\fontsize{10}{12}\selectfont 
\caption{Student's t-test MRR of COMPGCN and RGCN w./w.o. SymCL for link prediction. Note that all of the p-value $\textless$ 0.05.}
\resizebox{\linewidth}{!}{
\begin{tabular}{cccccc}
\hline
\textbf{Dataset}           & \textbf{Method} & \textbf{Run 1} & \textbf{Run 2} & \textbf{Run 3} & \textbf{p-value}        \\\hline
\multirow{4}{*}{WN18RR}    & COMPGCN         & 0.469          & 0.467          & 0.468          & \multirow{2}{*}{0.0008} \\
& COMPGCN-SymCL   & 0.471          & 0.471          & 0.472          &                         \\
& RGCN           & 0.429          & 0.426          & 0.427            & \multirow{2}{*}{0.0012}       \\
& RGCN-SymCL     & 0.432          & 0.431          & 0.431              &                         \\ \hline
\multirow{4}{*}{FB15k-237} & COMPGCN         & 0.351          & 0.351          & 0.353          & \multirow{2}{*}{0.0009} \\
& COMPGCN-SymCL   & 0.354          & 0.354          & 0.355          &                         \\
& RGCN           & 0.248          & 0.247          & 0.248               & \multirow{2}{*}{0.0377}       \\
& RGCN-SymCL     & 0.250          & 0.249          & 0.249         &                     \\\hline
\multirow{4}{*}{NELL-995}  & COMPGCN         & 0.456          & 0.458          & 0.452          & \multirow{2}{*}{0.0021} \\
& COMPGCN-SymCL   & 0.469          & 0.470          & 0.465          &                         \\
& RGCN             & 0.382          & 0.383          & 0.381             & \multirow{2}{*}{0.0008}       \\
& RGCN-SymCL      & 0.394          & 0.395          & 0.392               &                         \\ \hline
\end{tabular}}
\label{ttestR}
\vspace{-0.3 cm}
\end{table}

\begin{table*}[!t]
\fontsize{8}{11}\selectfont 
\caption{Performance comparison of \sysname{} with KGE baselines for link prediction. Boldface values indicate the best results.}
\resizebox{\linewidth}{!}{
\begin{tabular}{ccccccccccccc}
\hline
\multicolumn{1}{c}{\multirow{2.5}{*}{Methods}} & \multicolumn{4}{c}{\multirow{1}{*}{WN18RR}}                                                                                                                & \multicolumn{4}{c}{\multirow{1}{*}{FB15K-237}}                                                                                                                             & \multicolumn{4}{c}{\multirow{1}{*}{NELL-995}}                                                                                                     \\
 \cline{2-13} 
\multicolumn{1}{c}{}                         & \multirow{1}{*}{MRR}               & \multirow{1}{*}{Hit@1}             & \multirow{1}{*}{Hit@3}             & \multicolumn{1}{c}{\multirow{1}{*}{Hit@10}} & \multirow{1}{*}{MRR}               & \multicolumn{1}{c}{\multirow{1}{*}{Hit@1}} & \multicolumn{1}{c}{\multirow{1}{*}{Hit@3}} & \multicolumn{1}{c}{\multirow{1}{*}{Hit@10}} & \multirow{1}{*}{MRR}               & \multirow{1}{*}{Hit@1}             & \multirow{1}{*}{Hit@3}             & \multirow{1}{*}{Hit@10}                                             \\ \hline
\multicolumn{13}{c}{\textit{\textbf{Translational Distance Models}}}                                                                                                                                                                                       \\ \hline
\multicolumn{1}{c}{TransE}                   & 0.231                & 0.021                  & 0.409                  & \multicolumn{1}{c}{0.533}                   & 0.289                     & 0.193                     & 0.326                     & \multicolumn{1}{c}{0.478}                   & 0.249                & 0.095                  & 0.377                  & 0.471                   \\
\multicolumn{1}{c}{RotatE}                   & 0.476                & 0.428                  & 0.492                  & \multicolumn{1}{c}{0.571}                   & 0.338                     & 0.241                     & 0.375                     & \multicolumn{1}{c}{0.533}                   & 0.418                  & 0.313                    & 0.468                    & 0.620                     \\
\multicolumn{1}{c}{QuatE}                    & 0.481                & 0.436                  & 0.500                    & \multicolumn{1}{c}{0.564}                   & 0.311                     & 0.221                     & 0.342                     & \multicolumn{1}{c}{0.495}                   &   0.390             &  0.292                      &    0.433                    &  0.583                       \\
\multicolumn{1}{c}{DualE}                    & 0.482                & 0.440                   & 0.500                    & \multicolumn{1}{c}{0.561}                   & 0.330                     & 0.237                     & 0.363                     & \multicolumn{1}{c}{0.518}                   &   0.400                   &   0.299                        &         0.451                  &     0.595                       \\
\multicolumn{1}{c}{HAKE}                     & {\textbf{0.497}}       & {0.453}         & {\textbf{0.515}}         & \multicolumn{1}{c}{0.582}                   & 0.335                     & 0.237                     & 0.371                     & \multicolumn{1}{c}{0.530}                    & 0.415                & 0.313                  & 0.464                  & 0.612                   \\ \hline
\multicolumn{13}{c}{\textit{\textbf{Semantic Matching Models}}}                                                                                                                             \\ \hline
\multicolumn{1}{c}{RESCAL}                   & 0.455                & 0.419                  & 0.461                  & \multicolumn{1}{c}{0.493}                   & \multicolumn{1}{c}{0.353} & \multicolumn{1}{c}{0.264} & \multicolumn{1}{c}{0.385} & \multicolumn{1}{c}{0.528}                   &     0.176                & 0.078                  & 0.173                  & 0.314                \\
\multicolumn{1}{c}{DisMult}                  & 0.420                & 0.370                  & 0.439                  & \multicolumn{1}{c}{0.521}                  & 0.243                     & 0.191                     & 0.271                     & \multicolumn{1}{c}{0.328}                   & 0.223                & 0.139                  & 0.244                  & 0.398                   \\
\multicolumn{1}{c}{ComplEX}                  & 0.440                & 0.410                  & 0.460                  & \multicolumn{1}{c}{0.510}                   & \multicolumn{1}{c}{0.346} & \multicolumn{1}{c}{0.256} & \multicolumn{1}{c}{0.386} & \multicolumn{1}{c}{0.525}                   & 0.246                & 0.159                  & 0.281                  & 0.423                   \\
\multicolumn{1}{c}{ConvE}                    & 0.430                & 0.400                  & 0.440                  & \multicolumn{1}{c}{0.520}                   & \multicolumn{1}{c}{0.325} & \multicolumn{1}{c}{0.237} & \multicolumn{1}{c}{0.356} & \multicolumn{1}{c}{0.501}                             &    0.296                  &      0.226                  &    0.331                    &  0.430                \\
\multicolumn{1}{c}{ComplEX-DURA}             & 0.489                & 0.445                  & 0.503                  & \multicolumn{1}{c}{0.574}                   & \multicolumn{1}{c}{0.370} & \multicolumn{1}{c}{0.275} & \multicolumn{1}{c}{0.409} & \multicolumn{1}{c}{0.562}                   & 0.468                & 0.375                  & 0.513                  & 0.645                   \\ \hline
\multicolumn{13}{c}{\textit{\textbf{GNN-based Models}}}                                                                                                                              \\\hline 
\multicolumn{1}{c}{RGCN}                     & 0.427                & 0.382                  & 0.446                  & \multicolumn{1}{c}{0.510}                   & 0.248                     & 0.153                     & 0.258                     & \multicolumn{1}{c}{0.414}                   & 0.382                & 0.272                  & 0.435                  & 0.59                    \\
\multicolumn{1}{c}{SCAN}                     & 0.470                & 0.430                  & 0.480                  & \multicolumn{1}{c}{0.540}                   & \multicolumn{1}{c}{0.350} & \multicolumn{1}{c}{0.260} & \multicolumn{1}{c}{0.390} & \multicolumn{1}{c}{0.540}                   & --                     &  --                         &          --                 &  --                          \\
\multicolumn{1}{c}{KBGAT}                    & 0.464                & 0.426                  & 0.479                  & \multicolumn{1}{c}{0.539}                   & \multicolumn{1}{c}{0.350} & \multicolumn{1}{c}{0.260} & \multicolumn{1}{c}{0.385} & \multicolumn{1}{c}{0.531}                   & 0.377                & 0.299                  & 0.314                  & 0.430                   \\
\multicolumn{1}{c}{COMPGCN}                  & 0.469                & 0.434                  & 0.482                  & \multicolumn{1}{c}{0.537}                   & 0.352                     & 0.261                     & 0.387                     & \multicolumn{1}{c}{0.534}                   & 0.456                & 0.361                  & 0.507                  & 0.637                   \\ \hline
\multicolumn{13}{c}{\textit{\textbf{Ours Proposed Methods}}}                                                                   \\ \hline
\multicolumn{1}{c}{ComplEX-DURA-SymCL}       & 0.491                & 0.448                  & 0.504                 & \multicolumn{1}{c}{0.576}                   & \textbf{0.371}            & \textbf{0.276}            & \textbf{0.411}            & \multicolumn{1}{c}{\textbf{0.566}}          & 0.469                & 0.376                  & 0.511                  & 0.645                   \\
\multicolumn{1}{c}{HAKE-SymCL}               & \multicolumn{1}{c}{\textbf{0.497}}          & \multicolumn{1}{c}{\textbf{0.454}}          & \multicolumn{1}{c}{\textbf{0.515}}       & \multicolumn{1}{c}{\textbf{0.585}}             & 0.346                     & 0.248                     & 0.384                     & \multicolumn{1}{c}{0.544}                   & 0.419                & 0.318                  & 0.468                  & 0.616                   \\
\multicolumn{1}{c}{COMPGCN-SymCL}            & 0.471                & 0.437                  & 0.484                  & \multicolumn{1}{c}{0.537}                   & 0.354                     & 0.262                     & 0.389                     & \multicolumn{1}{c}{0.537}                   & \textbf{0.469}       & \textbf{0.378}         & \textbf{0.520}          & \textbf{0.649}          \\ \hline
\end{tabular}}
\label{LP_performance_table}
\end{table*}
\begin{table}[!h]
\fontsize{6}{8}\selectfont 
\caption{Performance comparison between \sysname{} with KGE baselines for entity classification. The boldface values indicate the best results.}
\resizebox{\linewidth}{!}{
\begin{tabular}{ccccc}
\hline
{Methods}        & {AIFB}           & {MUTAG}          & {BGS}            & {AM}             \\\midrule[0.25pt]
\multirow{1}{*}{Feat}         & \multirow{1}{*}{{55.55}} & \multirow{1}{*}{77.94}          & \multirow{1}{*}{72.41}          & \multirow{1}{*}{66.66}          \\
\multirow{1}{*}{WL}         & \multirow{1}{*}{{80.55}} & \multirow{1}{*}{80.88}          & \multirow{1}{*}{86.20}          & \multirow{1}{*}{87.37}          \\
\multirow{1}{*}{RDF2Vec}         & \multirow{1}{*}{{88.88}} & \multirow{1}{*}{72.06}          & \multirow{1}{*}{{{86.21}}}     & \multirow{1}{*}{87.88}          \\
\multirow{1}{*}{RGCN}       & \multirow{1}{*}{{95.83}}          & \multirow{1}{*}{72.21}          & \multirow{1}{*}{81.38}          & \multirow{1}{*}{89.19}         
 \\ 
\multirow{1}{*}{COMPGCN}   & \multirow{1}{*}{{94.44}} & \multirow{1}{*}{{79.29}} & \multirow{1}{*}{{82.35}} & \multirow{1}{*}{{{93.10}}}\\
\multirow{1}{*}{RR-GCN}   & \multirow{1}{*}{{83.33}} & \multirow{1}{*}{\textbf{81.67}} & \multirow{1}{*}{{80.00}} & \multirow{1}{*}{{70.00}}\\
\hline
\multirow{1}{*}{RGCN-SymCL} & \multirow{1}{*}{\textbf{96.11}} & \multirow{1}{*}{{72.35}} & \multirow{1}{*}{{83.45}} & \multirow{1}{*}{{89.60}}
\\
\multirow{1}{*}{COMPGCN-SymCL}   & \multirow{1}{*}{{94.44}} & \multirow{1}{*}{{{80.90}}} & \multirow{1}{*}{\textbf{88.24}} & \multirow{1}{*}{\textbf{96.55}}
     \\ \hline
\end{tabular}}
\vspace{-0.2 cm}
\label{EEEE_table}
\end{table}

\subsubsection{Superiority of \sysname{}}
We also compare \sysname{} with other state-of-the-art (SOTA) KGE models for both link prediction and entity classification tasks. The performance results are present in Tab. \ref{LP_performance_table} and Tab. \ref{EEEE_table}. Tab. \ref{LP_performance_table} indicates that \sysname{} outperforms other KGE models on all of the datasets, especially for the FB15k-237 and NELL-995. Compared to these two datasets, WN18RR is more simple and small. Thus, the basic KGE models are enough to get promising performance without contrastive learning frameworks, which leads to fewer improvements. Tab. \ref{EEEE_table} shows that \sysname{} makes an average 1.25\% improvement in accuracy compared to the previous SOTA KGE models for entity classification tasks. In particular, the COMPGCN-SymCL model achieves better results on the BGS and AM datasets. The experiment results indicate that our SymCL enhances the basic KGE encoders for better expressive and discriminative ability.

\subsubsection{Discussion}
The performance comparison on the two downstream tasks between \sysname{} with other KGE baselines shown in this section demonstrates the generalizability and superiority of our model. In particular, the generalizability, as the most important attribute of \sysname{}, is verified from two aspects: \textbf{(1)} \sysname{} can be easily adopted to enhance the expressive ability of different KGE models. \textbf{(2)} \sysname{} is applied to various benchmark datasets for different downstream tasks. Moreover, the promising results suggest that the structural information leveraged in \sysname{} indeed helps model learn more powerful and discriminative representations.

\subsection{Relation-Symmetrical Structure Analysis}
\subsubsection{Statistical Analysis}
We counted and recorded the number and calculated the proportion of the \emph{1-hop} and \emph{2-hop} relation-symmetrical structures in three link prediction benchmark datasets to demonstrate the universality of the defined relation-symmetrical structures. {During relation-symmetrical structure extraction, we simply remove the recurring nodes in our realization, which would scale to the dataset with more relations to alleviate the potential over-representation problem.} Fig. \ref{beta3} suggests that there are many relation-symmetrical (R-S) structures in these datasets, which suggests the feasibility of the motivation in this work. In particular, compared to WN18RR and FB15K-237, more R-S structures are found in NELL-995, which may also be a reason for the more apparent improvements made by \sysname{} on NELL-995 datasets.
\begin{figure*}[t]
\centering
\includegraphics[width=\textwidth]{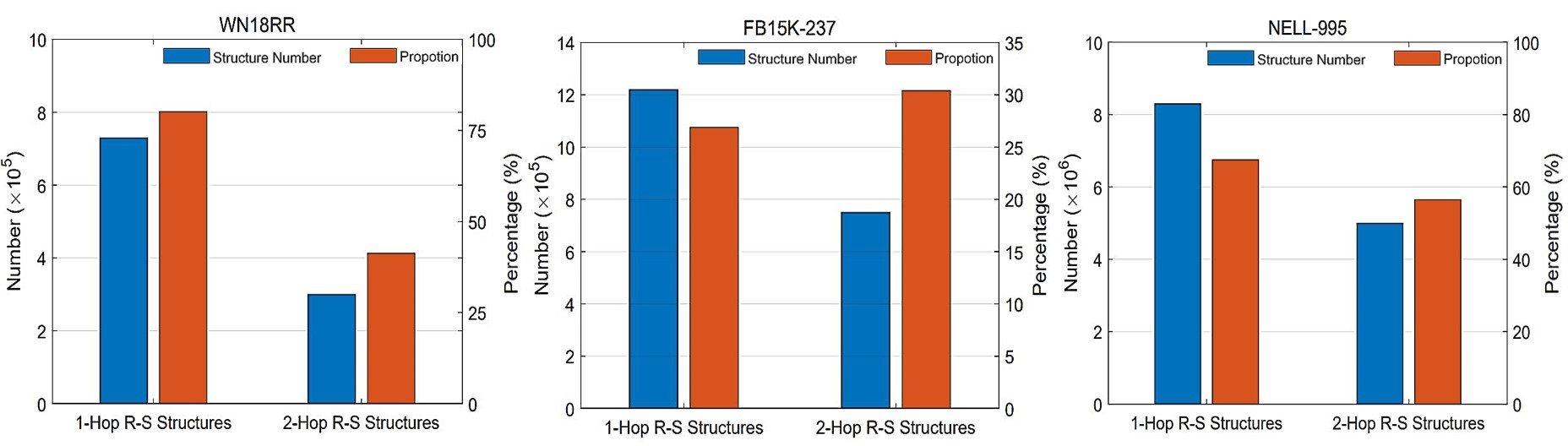} 
\caption{Statistics on relation-symmetrical (R-S) structures in benchmark datasets. The left Y axes correspond to the structure number bars, \textit{i.e.,} blue bars, while the right Y axes correspond to the proportion bars, \textit{i.e.,}  red bars. {Proportion is calculated by "\#k-hop R-S structures/\#k-hop structures".}}
\label{beta2}
\vspace{-0.2 cm}
\end{figure*}

\begin{figure}[t]
\centering
\includegraphics[width=0.48\textwidth]{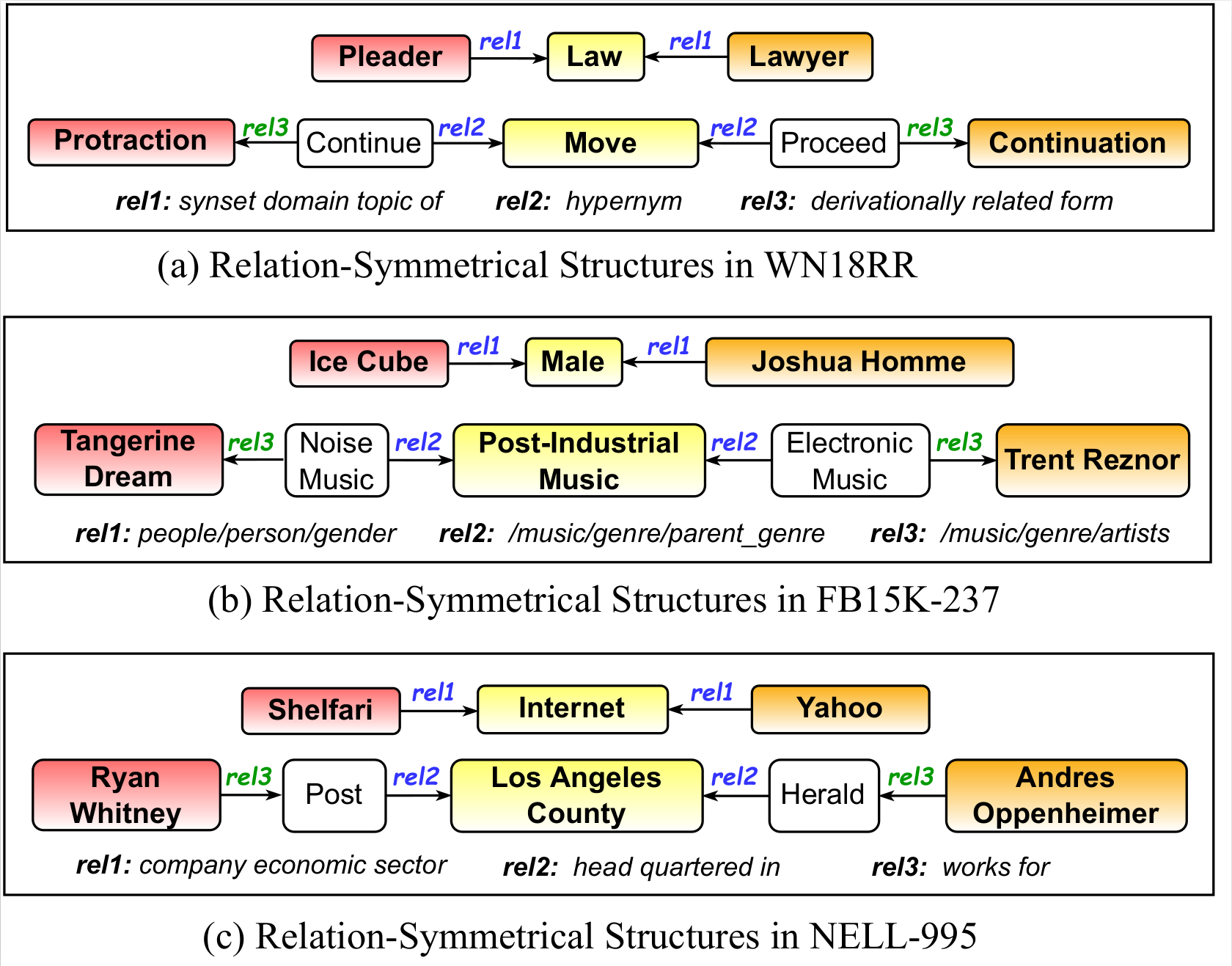}
\vspace{-0.2 cm}
\caption{Relation-Symmetrical structure samples}
\label{beta4}
\vspace{-0.4 cm}
\end{figure}

\subsubsection{Intuitive Case Study}
We further intuitively show the six relation-symmetrical structures from the link prediction benchmark datasets, \textit{i.e.,} WN18RR \cite{WN18RR}, FB15K-237 \cite{FB15K-237}, and NELL-995 \cite{NELL-995} (See Fig. \ref{beta4}). For each dataset, a \emph{1-hop} relation-symmetrical structure and a \emph{2-hop} relation-symmetrical structure are present. These cases demonstrate that entities located in relation-symmetrical positions, commonly found in knowledge graphs, are usually similar and correlated. For example, \emph{Tangerine Dream} is \emph{2-hop} relation-symmetrical to \emph{Trent Reznor} about \emph{Post-Industrial Music}. Meanwhile, we can easily find that these entities have similar semantics since they are both \emph{Post-Industrial Music} artists in Fig. \ref{beta4} (b). Besides, \emph{Shelfari} is \emph{1-hop} relation-symmetrical to \emph{Yahoo} about \emph{Internet}, which indicates the semantic similarity between them since they are both \emph{Internet} companies in Fig. \ref{beta4} (c). Similar conclusions also can be deducted from other samples. Note that we do not need to know the concrete label of the entity or the relationships in our relation-symmetrical structure. We can just number the relationships as \emph{re11}, \emph{re12}, \emph{re13} and so on. If the same relation types associated with the edge directions are symmetrical in the structure, we can treat it as the relation-symmetrical structure referred to in Def. 2. It is because the relational edges in KGs naturally constrain the entities. Starting from the same pivot entity (\textit{e.g.,} Male) and the same \emph{rel1}, we will reach the entities with similar semantics, no matter what the pivot entity and \emph{rel1} are. Using such property of the relation-symmetrical structure to construct the contrastive pairs will be more stable since no other language models are required.

\subsection{Running Time Analysis}
\begin{figure}[t]
\centering
\includegraphics[width=0.48\textwidth]{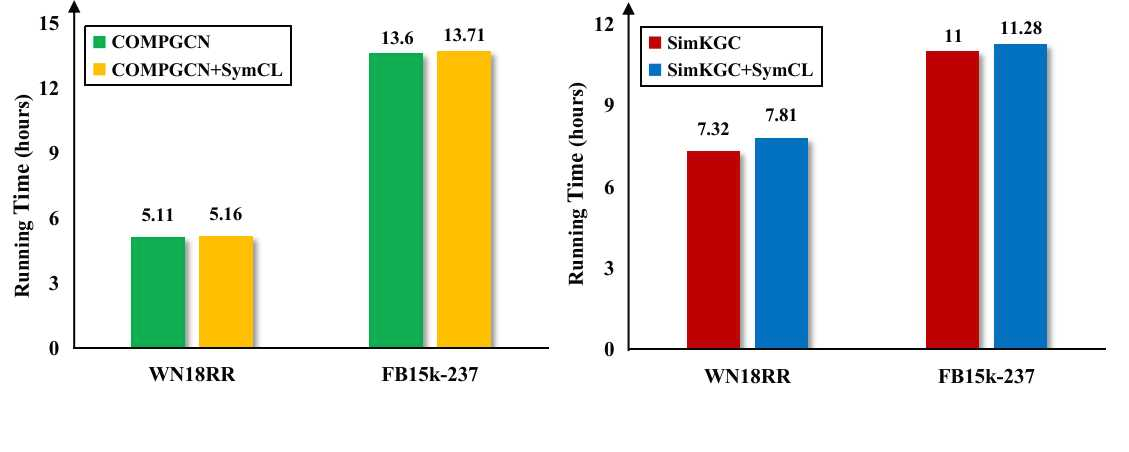}
\vspace{-0.5 cm}
\caption{{Running time comparison of SimKGC and COMPGCN w./w.o SymCL.}}
\label{time}
\end{figure}

We analyze the running time of our method in this section. Our method contains two parts, \textit{i.e.,} relation-symmetrical structure extraction and relation-symmetrical contrastive learning. As first contrastive KGE framework based on structural information, We focus more on the effectiveness of \sysname{}. Thus, we only leverage the traversing method to extracting the relation-symmetrical structures and store them into the corresponding dictionary. We observe that the computational costs are acceptable for all the typical KG benchmark datasets since we only need to run the structure extraction once for each dataset and reuse the restored dictionary $npy$ files, which will be shared after the paper is published. More concretely, it takes about {50}, {120}, and {95} mins with intel core i9-9900K CPU to extract the \emph{2-hop} structures for WN18RR, FB15K-237, and NELL-995. Besides, we further compare the running time of the learning procedure w./w.o our relation-symmetrical contrastive framework. {Fig. \ref{time} indicates that the running time caused by our contrastive learning model is only 2.5 and 6.7 mins for WN18RR and FB15k-237 datasets with COMPGCN \cite{COMPGCN}, respectively. Moreover, it is found that the running time is increased by only 29 mins and 16 mins for WN18RR and FB15k-237 datasets with SimKGC \cite{simkgc}, which is acceptable and shows great opportunity to extend our \sysname{} with LLM. Considering the performance improvement, These small proportions of the time increase are acceptable.}

\subsection{Hyper-parameter Analysis}
\begin{figure}[t]
\centering
\includegraphics[width=0.48\textwidth]{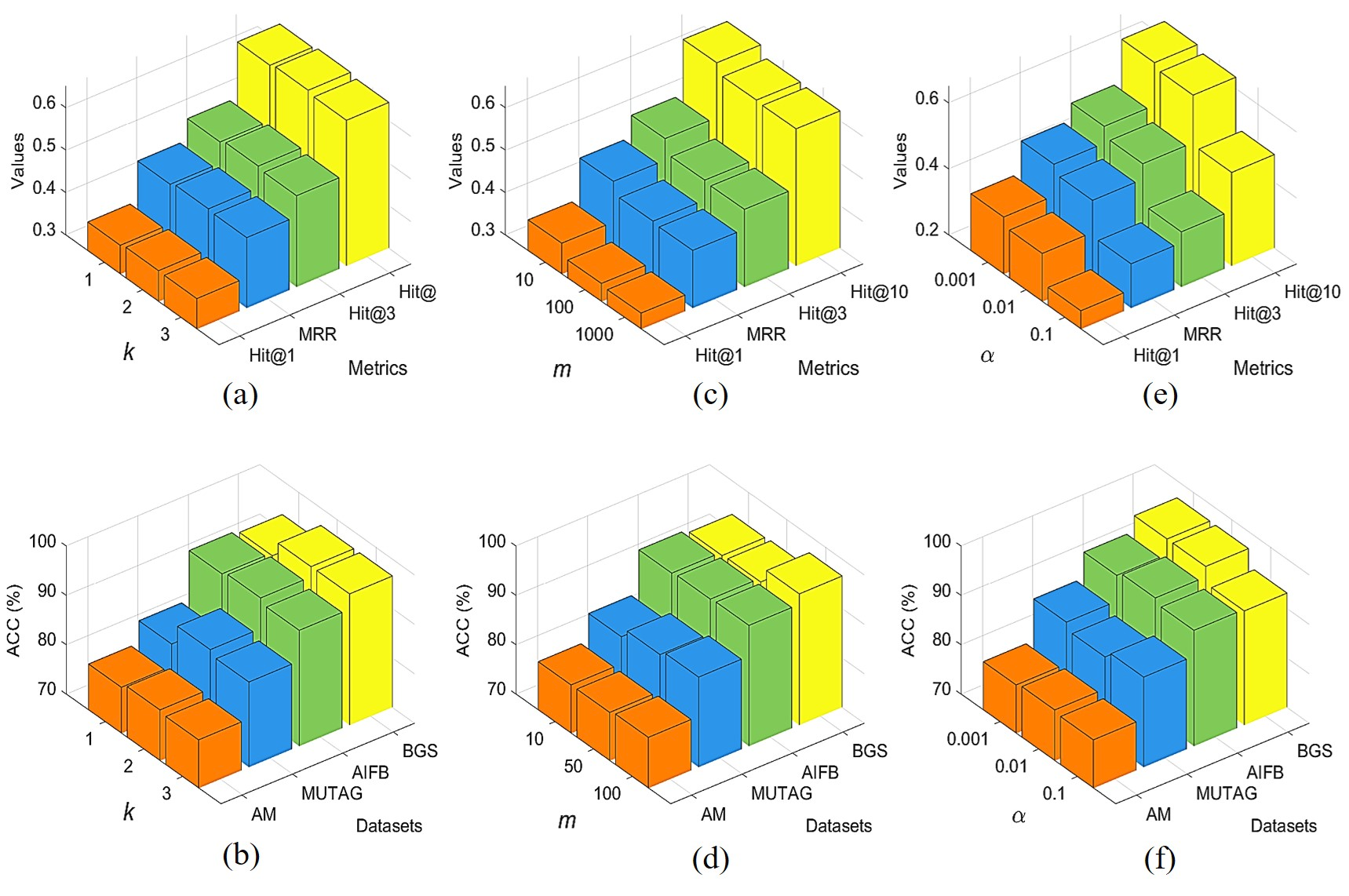} 
\caption{Sensitivity analysis of the hyper-parameters. Sub-figure (a), (c), and (e) correspond to link prediction, and the rest parts are related to entity classification. }
\label{beta3}
\end{figure}

We investigate the influence of the hyper-parameter hop $K$, sampling number $m$, and trade-off weight $\alpha$ in our \sysname{}. The COMPGCN-SymCL is selected as the basic model. As for the scope of the hyper-parameters for both two tasks, \textit{i.e.,} entity classification and link prediction, $K$ is selected in $\{1, 2, 3\}$, and $\alpha$ is searched in $\{0.001, 0.01, 0.1 \}$. However, since the benchmark datasets for entity classification are smaller than link prediction, the different scopes of $m$ are used, i.e., $\{10, 50, 100, 1000\}$ for link prediction and $\{10, 50, 100\}$ for entity classification. As for $K$ and $m$, it is observed that there is no great fluctuation of performance when $K$ and $m$ are varying in Fig. {\ref{beta3}} (a) to {\ref{beta3}} (d). It demonstrates that \sysname{} is insensitive to $K$ and $m$. The reason is that there are many symmetrical structures for each entity, which can enhance the discriminative capability of samples. As for the trade-off hyper-parameter $\alpha$, we find our \sysname{} is much more sensitive to it (See Fig. {\ref{beta3}} (e) and (f)). It is because of the magnitude difference between the contrastive and task loss, i.e., the contrastive loss is usually a hundred times larger than the task loss. The best performances are generally reached when $\alpha=0.001$.

\section{Conclusion}
In this paper, we propose a knowledge graph contrastive learning framework based on relation-symmetrical structures, termed \sysname{}, which leverages the symmetrical structural information in KGs to enhance the discriminative ability of KGE models. Extensive Experimental results on benchmark datasets have verified the proposed contrastive framework can be easily adopted to other KGE models to enhance their discriminative and expressive ability. {Although proven effective, we can definitely further improve the discriminative capacity of \sysname{} if high-confidence negative contrastive pairs can be constructed and leveraged. As a primitive attempt to use structure information for contrastive learning KGE, it is not studied in this work.} In the future, we aim to continue investigating a more fine-grained strategy for high-confidence negative pair construction to empower our contrastive learning framework, such as integrating our \sysname{} with other language-based contrastive KGE models.



\ifCLASSOPTIONcompsoc
  \section*{Acknowledgments}
\else
  \section*{Acknowledgment}
\fi

This work was supported by the National Key R$\&$D Program of China (project no. 2020AAA0107100) and the National Natural Science Foundation of China (project no. 62276271). 

\ifCLASSOPTIONcaptionsoff
  \newpage
\fi



\bibliographystyle{IEEEtran}
\bibliography{2_reference}




\end{document}